\documentclass{article}
\usepackage[utf8]{inputenc}
\usepackage{indentfirst}
\usepackage{hyperref}
\usepackage{mathtools}
\usepackage{graphicx,kantlipsum}
\usepackage{tikz,graphics,color,fullpage,float,epsf,caption,subcaption}
\usepackage{authblk}
\usepackage[intoc]{nomencl}
\usepackage{setspace}
\usepackage{amsmath,amssymb}
\DeclareMathOperator{\E}{\mathbb{E}}
\usepackage[a4paper, total={6in, 8in}]{geometry}
\usepackage{hyperref}
\hypersetup{
    colorlinks=true,
    linkcolor=blue,
    filecolor=magenta,      
    urlcolor=cyan,
    pdftitle={Overleaf Example},
    pdfpagemode=FullScreen,
    }

\graphicspath{ {./images/} }
\title{\bf ApolloRL: a Reinforcement Learning Platform for Autonomous Driving}
\author[ ]{Fei Gao\thanks{
        {Correspondence Author: Fei Gao (\tt\small gaofei09@baidu.com)}}, Peng Geng, Jiaqi Guo\thanks{
        {Work done while interning at Baidu Apollo. Now with University of Cambridge.}}, Yuan Liu, Dingfeng Guo, Yabo Su, Jie Zhou, Xiao Wei, Jin Li, Xu Liu}
\affil[ ]{Baidu Apollo Autonomous Driving}
\date{\vspace{-5ex}}

\begin{document}

\maketitle
\begin{abstract}
    We introduce \textit{ApolloRL} \footnote{Platform link:  \url{https://studio.apollo.auto/pnc-platform}}, an open platform for research in reinforcement learning for autonomous driving. The platform provides a complete closed-loop pipeline with training, simulation, and evaluation components. It comes with 300 hours of real-world data in driving scenarios and popular baselines such as Proximal Policy Optimization (PPO) and Soft Actor-Critic (SAC) agents. We elaborate in this paper on the architecture and the environment defined in the platform. In addition, we discuss the performance of  the baseline agents in the ApolloRL environment.

\end{abstract}

\section{Introduction}
\begin{figure}[h]
    \centering
    \includegraphics[width=0.8\textwidth]{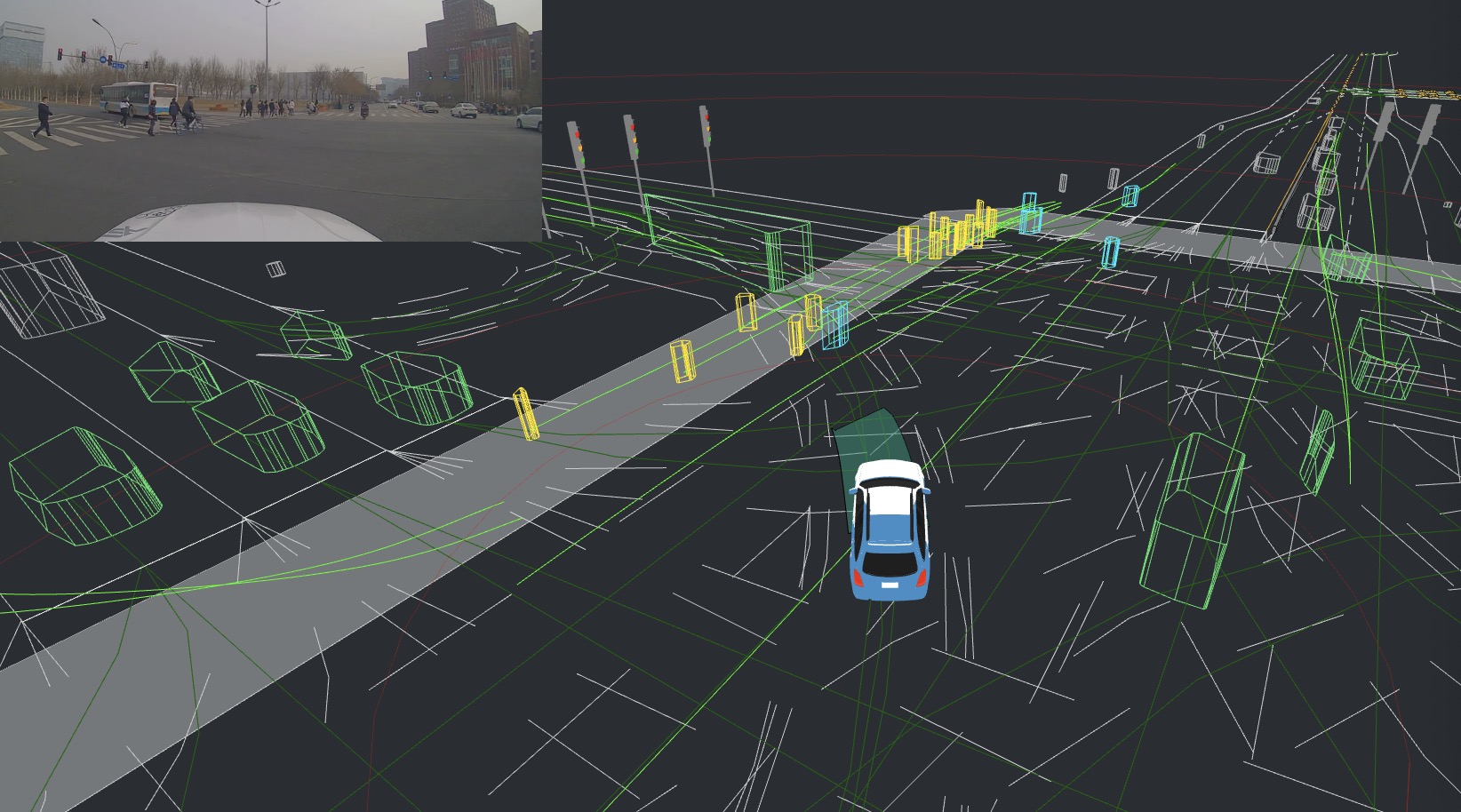}
    \caption{The scenarios in simulation}
    \label{fig:env}
\end{figure}
Autonomous driving \cite{av} has received much attention due to its potential huge impact to our world, \textit{e.g.}, increased driving safety \cite{av-safety}, improved transportation efficiency \cite{Hancock7684}, and reduced commuting time \cite{Steck2018HowAD}. The autonomous driving technology has advanced significantly over the past few years, thanks to the research progress in artificial intelligence. However, the autonomous driving problem still remains  challenging in multiple aspects. For example,
\begin{itemize}
    \item \textit{Imperfect information}: Autonomous driving is an imperfect information problem. There are noises, defects, restrictions of sensors or occlusion in perception and localization. It is also difficult to precisely know the intentions of other traffic participants.
    \item \textit{Multiple objectives}: It is hard to define a single metric for autonomous driving. We need to consider multiple requirements all at once, such as safety, passenger comfort, reachability, efficiency, traffic regulations, and sociality.
    \item \textit{Complex action space}: The driving scenarios are complex and diverse (\textit{e.g.,} see Fig.~\ref{fig:env}). Despite the low dimensionality, the action space is continuous and combinatorial, and it must comply with the constraints in dynamics.
    \item \textit{Delayed rewards}: The driving process is long, and the impact on the future needs to be considered. For example, the impact time is around 5-10 seconds (50-100 frames), which can be even longer on the lane change strategy. 
\end{itemize}

The problem of the autonomous driving decision-making and planning can be regarded as a real-world, open-ended, complex and strictly constrained game. Although there have been many autonomous systems using machine learning, deep learning, traditional robotics, numerical methods, and rule-based methods, they do not seem well address the above-mentioned problems. Reinforcement learning, a learning paradigm that has made great success in games and robotics \cite{SilverHuangEtAl16nature,dota2,muzero,openai2019learning,mirowski2018learning,streetair_iccv2019}, seems a promising direction to solve these challenges.  We believe a unified platform could benefit the research community of autonomous driving and reinforcement learning, by providing a comprehensive engineering architecture, strong computing power and large scale real-world data.

\begin{figure}[t]
    \centering
    \includegraphics[width=\textwidth]{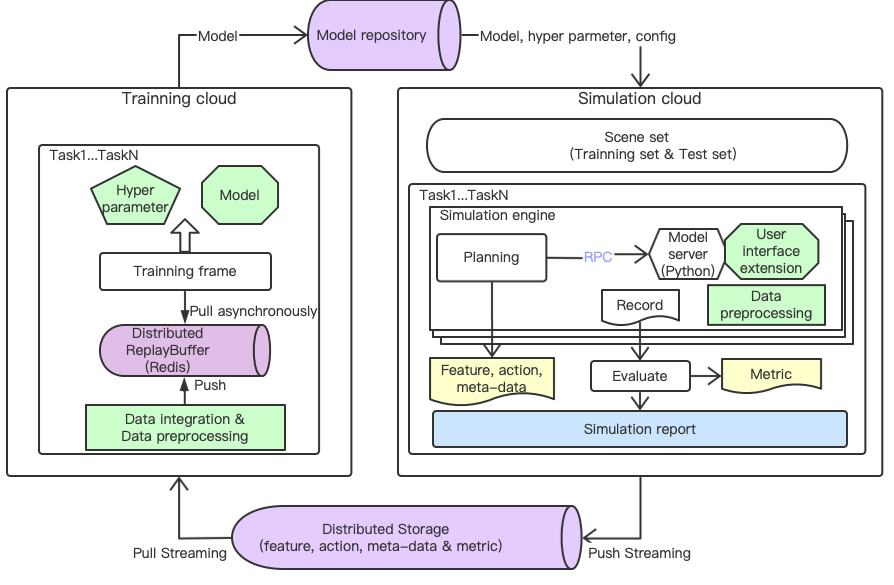}
    \caption{ApolloRL platform architecture}
    \label{fig:structure}
\end{figure}

This paper introduces the first open motion planning reinforcement learning platform, named \textit{ApolloRL}, for autonomous driving with integrated model training and simulation evaluation. 
The overall architecture of the ApolloRL platform is shown in Fig. \ref{fig:structure}. The platform has multiple advantages such as real-world data from on-road logs, powerful functions, comprehensive evaluation standards, and scalable architecture as detailed described below:
\begin{itemize}
    \item \textit{Data}: We provide a large number of simulation scenes from  complex real-world on-road logs with object interaction.
\item \textit{Function}: We provide a closed-loop pipeline from training to evaluation, allowing for quick iterations of the agent algorithms and performances.
\item \textit{Evaluation}: We provide a comprehensive and practical driving evaluation in traffic regulations, safety, comfort, intelligence, etc.
\item \textit{Engineering architecture}: A distributed simulation and a distributed trainer are implemented to reduce the effort of research engineering.
\item \textit{Extensibility}: We provide extensible interfaces for different needs of reinforcement learning. Developers can easily adjust parameters and implement their own RL algorithms.
\end{itemize}

Currently, the platform brings 300 hours of Autonomous Driving Scenario (ADS) from real-world data. We also support the mainstream continuous control reinforcement learning algorithms PPO and SAC. ApolloRL provides the two baseline agents, with the flexibility of using arbitrary hyperparameters, models, and exploration algorithms. We hope the platform could inspire and help with the advances of reinforcement learning research in autonomous driving.

\section{Related Work}
A popular application of reinforcement learning is gaming: including arcade learning environment (ALE \cite{ale}), StarCraft \cite{starcraft} and DeepMind Lab \cite{deepmindlab}. Another application domain is robotics: OpenAI Gym such as MuJoCo Robotics \cite{openaigym}, etc. However, in the field of real-world tasks, especially in the field of autonomous driving, there is a lack of environments that can provide real, large-scale, and standardized task sets. There are racing simulators like TORCS \cite{torcs} and traffic simulators like CARLA \cite{carla}, SUMO \cite{sumo}, and SMARTS \cite{smarts}. However, they do not provide real-world scenes, lack of some semantic elements, or have limited scenarios. Recently, DriverGym \cite{driving} provides an offline reinforcement learning environment that enables simulating reactive agents using data-driven models learned from real-world data, and also provides a lot of scenes from real-world logs. Our platform is the first open platform that is able to provide large-scale simulation and training capabilities with many real-world scenes.
\section{The ApolloRL Autonomous Driving Environment}
\begin{figure}[t]
    \centering
    \includegraphics[width=\textwidth]{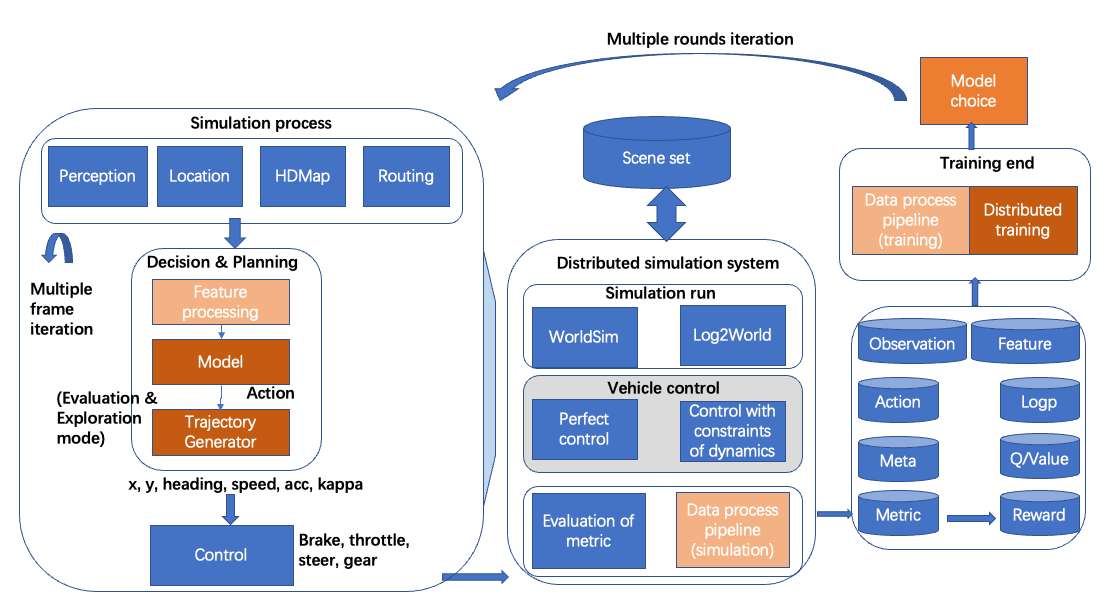}
    \caption{Overview of ApolloRL platform}
    \label{fig:structureflow}
\end{figure}
 The architecture and the data flow of ApolloRL are shown in Fig. \ref{fig:structureflow}. The input perception information is fed into the Planning \& Decision module for trajectory planning. The trajectory output is then passed into the control module, which gives detailed commands on the vehicle to control in either the \textit{WorldSim} or \textit{Log2World} simulation environments. The ApolloRL autonomous driving environment allows the community to test their own RL implementations of the planner in the autonomous driving scenarios generated from the real-world data. The users can implement arbitrary RL algorithmsnd define their own environment settings, including observations, actions, and shaped rewards. We now describe the environment in detail.

\subsection{Scenario Description}
Self-driving vehicles use sophisticated sensors to capture the surrounding environment. A total of about 300 hours of 60,000 ADSs from real-world driving data, including 50,000 for training and 10,000 for testing have been collected and made available to the community on the platform. These are then reproduced virtually in the Apollo environment, as shown in Fig. \ref{fig:env}. The various scenarios categorized by the map, consisting of other vehicles, railings, pedestrians, and non-motorized vehicles, have been reproduced from the real-world data, which are much more realistic than the simulated environments. 

According to the driving scene, the main complex scenarios have been split into short clips and divided into several categories, including going straight through the intersections, turning left/right through the intersections, going straight in the non-intersection road, etc. These will be referred to as \textit{scene/scene set} in this paper. For example, a left-turning scene starts at a certain distance prior to the intersection and ends at a certain distance after the left turn.

In an ADS, the ego vehicle starts from a fixed position, corresponding to the real position when the scenario data was collected. Then it moves following whatever algorithm the users provide. The task is marked done when it reaches its goal without encountering any collision. The simulation ends when any of these following situations happens: (1) when any collision or irrational behavior is detected; (2) when the planned movement does not follow dynamics constraints; (3) when the ego vehicle goes off the road; (4) when the original record ends. 

\subsection{Simulations}

The simulation environment, the metric evaluation standard, and the control algorithm used in the environment are consistent with those used in the real vehicle's system to ensure the portability. The controller then converts the trajectory of the first 0 $\sim$ (1-1.5) seconds into executable steering and acceleration. In addition, to ensure portability to the real autonomous driving situations, we used a dynamics simulation system instead of a perfect simulation (progressed as planned). Moreover, editing existing scenarios or even synthesizing scenarios are supported. The synthesized data is in the form of perturbations to the expert's driving trajectories, which creates rare and worse augmented data such as collisions or going off the road.

\subsection{Observations}
\begin{figure}[t]
    \centering
    \includegraphics[width=0.9\textwidth]{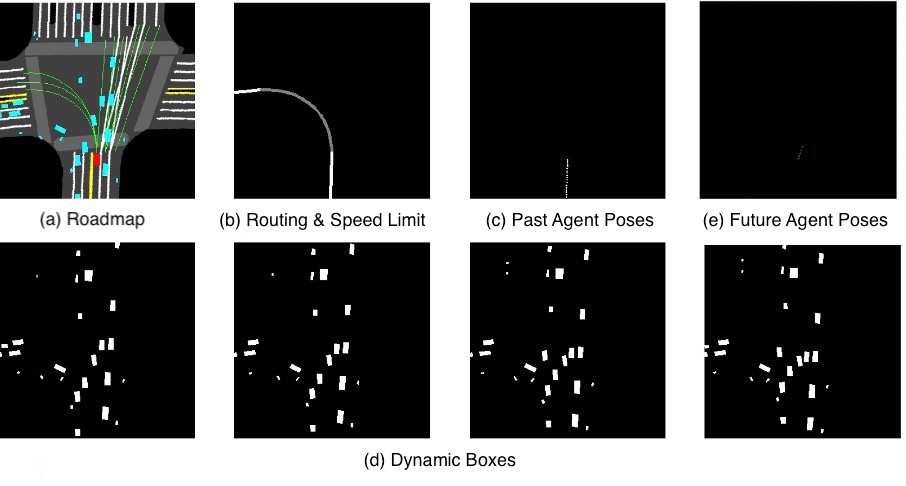}
    \caption{Observations (a-d) and output (e)}
    \label{fig:obs}
\end{figure}

Drawing on from the ChauffeurNet paper \cite{ChauffeurNet}, our perception system processes the raw camera and sensor data. The produced observation is a top-down representation of the environment and intended route, where objects such as vehicles are drawn as oriented 2D boxes along with a rendering of the road information and the traffic light states. At time \textit{t}, our agent is represented in a top-down coordinate system by $p_t$, $\theta_t$, $s_t$, where \textbf{$p_t$} = $(x_t, y_t)$ denotes the agent's location, $\theta_t$ denotes the heading, and $s_t$ denotes the speed. The top-down coordinate is picked so that our agent's location at the current time $t=0$ is always at a fixed location $(u_0, v_0)$ within the image. As shown in Figure \ref{fig:obs}, the observations consist of images of size $W \times H$ rendered into this top-down coordinate system. Several notes about the observations are discussed below:
\begin{itemize}
    \item[(a)] \textit{Roadmap}: a color (RGB) image with a rendering of various map features such as traffic lights, lanes, stop signs, crosswalks, curbs, etc. Within each frame, we color each lane center by a color representing red lights, yellow lights, or green lights. 
    \item[(b)] \textit{Routing and Speed limit}: a single channel (gray-scale) image with the lane centers colored in proportion to their known speed limit and the intended route along which we wish to drive, generated by a router. 
    \item[(c)] \textit{Past agent poses}: the past poses of our agents are rendered into a single gray-scale image as a trail of points. The sampling interval is fixed as $\delta t$ to sample past and future temporal information, such as the traffic light state or dynamic object states in the above inputs.
    \item[(d)] \textit{Dynamic objects in the environment}: a temporal sequence of images showing all the potential dynamic objects (vehicles, cyclists, pedestrians) rendered as oriented boxes \cite{ChauffeurNet}. 
\end{itemize}

The top-down 2D view allows for efficient convolutional input to the feature net of the RL agent and learnable spatial relationships. The difficulty of learning from the raw pixel input was revealed by Tian et al. \cite{pixel} and the advantage of intermediate representation was emphasized. 

\subsection{Actions}\label{sec:action}
The actions (output) of the agent is a planned driving trajectory $(x_1, y_1, \theta_1)$, $(x_2, y_2, \theta_2)$, $(x_3, y_3, \theta_3)$, ..., $(x_T, y_T, \theta_T)$ that would be consumed by a controller which then translates it to steering and acceleration, as shown in Fig. \ref{fig:obs}(e). The action space is the space covered by every possible future trajectory. Given a sample interval $\delta t$ and the number $n_{points}$ of points in a planned trajectory, we predict the next $T_{predict}=\delta t \times n_{points}$ time ahead. In our experiments, we take $\delta t = 0.2s$, $n_{points}=15$, which produce $T_{predict}=3s$.
\subsection{Reward Structure}
We define the basic extrinsic reward structure as follows: positive reward for reaching the goal of the scene, and negative rewards for collision, going off-road, going in the wrong direction, and the discomfort of the passenger. While the discomfort is ambiguous to define, the lateral, and the longitudinal accelerations serve as useful measurements. With the reward structure, the highest possible cumulative reward gained would be reaching the goal with little or no bad behaviors that introduce a negative reward. As the vehicle may not experience any reward while driving on the road, the basic structure could be viewed as a sparse reward. The community is welcome to design its own reward structure based on the internal reward, the metrics, and other information provided by the platform.

\section{Baseline Agents}
This section illustrates the baseline agents and demonstrates their capability in ApolloRL.
\subsection{Policy Learning}
The RL agent we use includes deep neural networks that predict the proper actions (trajectories) based on the observation (images) input. Drawing on from the ChauffeurNet \cite{ChauffeurNet}, we include a convolutional feature network (FeatureNet) that consumes the input data to generate feature representation that is shared by other networks. These features are then fed into the V/Q network and the Policy network. The RNN policy network outputs a trajectory distribution with parameters $\theta$, which defines a policy $\pi_{\theta}$. At time $t$ the agent receives observations $s_t$, selects an action $a_t$ with probability $\pi_\theta(a_t|s_t)$, and then receives a reward $r_t$ from the environment. The agent is trained to maximise the total return (cumulative reward) $G_t = \sum_{k=0}^\infty \gamma^k r_{t+k+1}$ where $\gamma$ is the discount factor. We implemented two state-of-the-art RL algorithms, PPO \cite{ppo} and SAC \cite{sac}. The implementation details are discussed below.
\begin{figure}[t]
    \centering
    \includegraphics[width=0.95\textwidth]{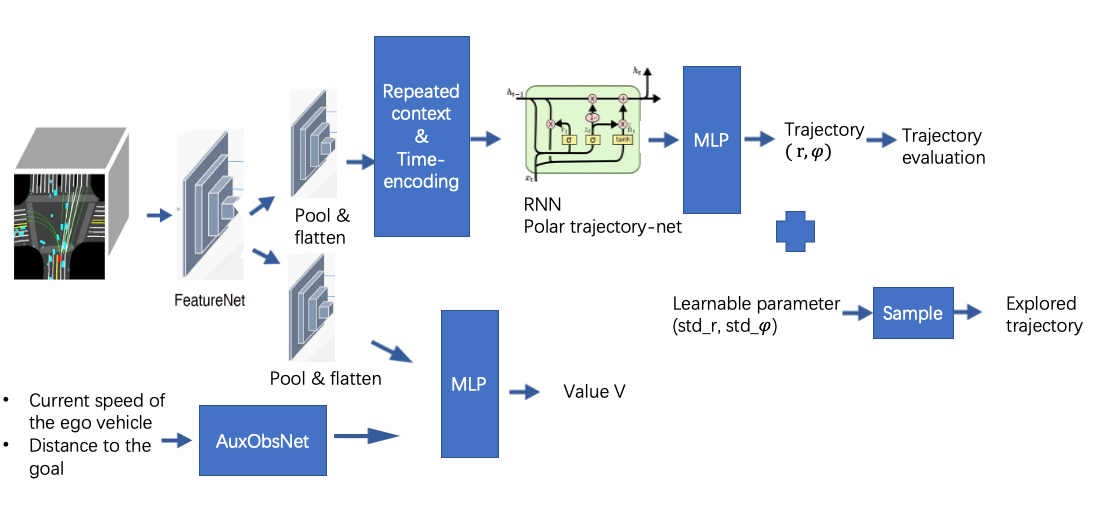}
    \caption{An illustration of the core model structure for a PPO agent, with a FeatureNet, a V-Net, a Pi-Net, and a learnable exploration std. parameter.}
    \label{fig:pponetwork}
\end{figure}
\subsubsection*{Proximal Policy Optimization (PPO)}
Proximal Policy
Optimization (PPO), as described by Schulman et al. \cite{ppo}, has demonstrated state-of-the-art performance while being much simpler to implement and tune, compared to other methods. The PPO loss is defined as follows:
\begin{equation}
    L_t^{CLIP+VF+S}(\theta) = \hat{E}_t[L_t^{CLIP}(\theta)-c_1 L_t^{VF}(\theta)+c_2 S[\pi_\theta](s_t) ]
\end{equation}
where $\theta$ is the policy parameter; $\hat{E}_t$ denotes the empirical expectation over timesteps; $L_t^{CLIP}$ is the clipped loss; $c1, c2$ are coefficients; $S$ denotes an entropy bonus; $L_t^{VF}$ is a squared-error loss $(V_\theta(s_t)-V_t^{targ})^2$). We use the PPO penalty variant which augments the clipped surrogate objective and makes use a learned state-value function $V(s)$, further combined with entropy bonus to ensure sufficient exploration. The mean of the policy output is used for evaluation, while the combined mean and the pre-defined exploration std. (standard deviation) are used for exploration. Then the explored trajectory is used for the next training session.
\subsubsection*{Soft Actor-Critic (SAC)}
\begin{figure}[h]
    \centering
    \includegraphics[width=0.95\textwidth]{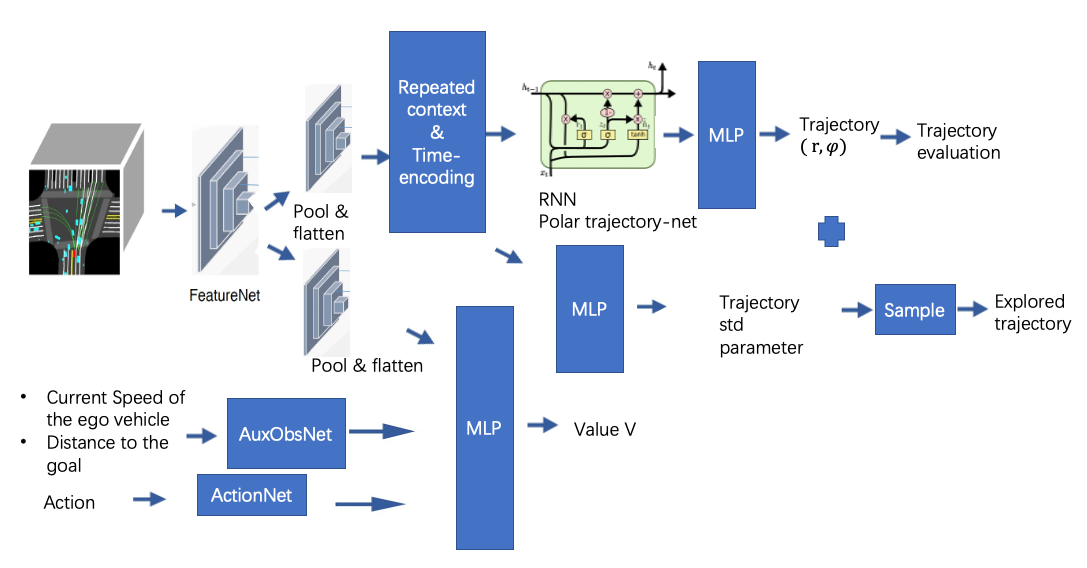}
    \caption{The core model structure for SAC, with a FeatureNet, two Q-Net, and a Pi-Net}
    \label{fig:sacnetwork}
\end{figure}
Soft Actor-Critic (SAC), as described by Haaroja et al. \cite{sac}, optimizes a stochastic policy in an off-policy manner. A central feature of SAC is entropy regularization. The policy is trained to maximize a trade-off between expected return and entropy, a measure of randomness in the policy. SAC learns a policy network and two Q-networks. We represent the log of the standard deviation (std) as the outputs from the neural network. The objective function is shown below.
\begin{equation} 
\pi_*=\arg\max_\pi \E_{(s_t,a_t)\sim \rho_\pi}[r(s_t,a_t)+\alpha H(\pi(\cdot | s_t))]
\end{equation}
where $\alpha$ is the temperature coefficient and $H(\cdot)$ is the entropy. There are three kinds of loss functions used in SAC algorithm: the loss of the Q network, the loss of the policy and the loss of the temperature coefficient. Their formulations are provided below.

\paragraph{The loss $J_{Q}(\phi)$ of the Q function.} The SAC agent contains two Q networks $Q(\cdot;\phi_1)$ and $Q(\cdot;\phi_2)$ where $\phi_i$ is their parameters. Their loss functions are defined as
\begin{equation}
    J_{Q}(\phi_i)=\E_{(s_t, a_t)\sim \rho_\pi}\left[ Q(s_t, a_t;\phi_i)-(r(s_t,a_t)+\gamma\E_{s_t\sim D}[V_{\Bar{\varphi}} (s_{t+1})]) \right],i \in (1,2)~.
\end{equation}
The value function $V_{\varphi}$ parameterized by $\varphi$ is defined as
\begin{equation}
    V_{\Bar{\varphi}}(s_{t+1})= \E_{(s_t,a_t) \sim \rho_\pi}\left[ \min_{i=1,2}\Bar{Q}(s_{t+1}, a_{t+1};\phi_i) - \alpha\log \pi_{\Bar{\theta}}(a_{t+1}|s_{t+1})\right],a_{t+1}\sim\pi_{\Bar{\theta}}(\cdot|s_{t+1})
\end{equation}
where $\min_{i=1,2}\Bar{Q}(s_{t+1}, a_{t+1};\phi_i)$ is the minimum of the two Q approximators. $\alpha$ is the (strictly positive) temperature coefficient.

\paragraph{The loss $J_\pi(\theta)$ of the policy.} The loss of the policy is defined as
\begin{equation}
J_\pi(\theta)=\E_{s_t\sim D}\left[
\E_{a_t \sim \pi_\theta}[\alpha \log\pi_\theta(a_t|s_t) - Q(s_t,a_t;\phi)]
\right] 
\end{equation}
where $D=\{(s_t, a_t, r(s_t, a_t), s_{t+1})\}$ is the set of episodes.

\paragraph{The loss $J(\alpha)$ of the temperature coefficient.}
\begin{equation}
    J(\alpha)=\E_{a_t \sim \pi_t}[\alpha \log\pi_\theta(a_t|s_t)-\alpha\Bar{H}]
\end{equation}
where $\Bar H$ is a pre-defined target entropy. 

\subsection{Exploration}
We discuss in this section the representation used for policy exploration.
\subsubsection{Cartesian Coordinate}
\begin{figure}[h]
     \centering
     \begin{subfigure}{1\textwidth}
         \centering
         \includegraphics[width=.9\linewidth]{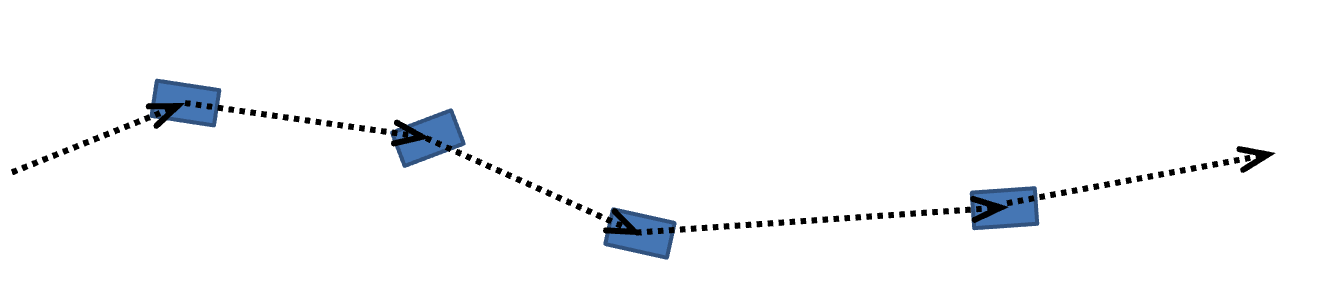}
         \caption{Actions output by the policy network}
         \label{fig:xy_explore0}
     \end{subfigure}
     \hfill
     \begin{subfigure}{1\linewidth}
         \centering
         \includegraphics[width=.9\textwidth]{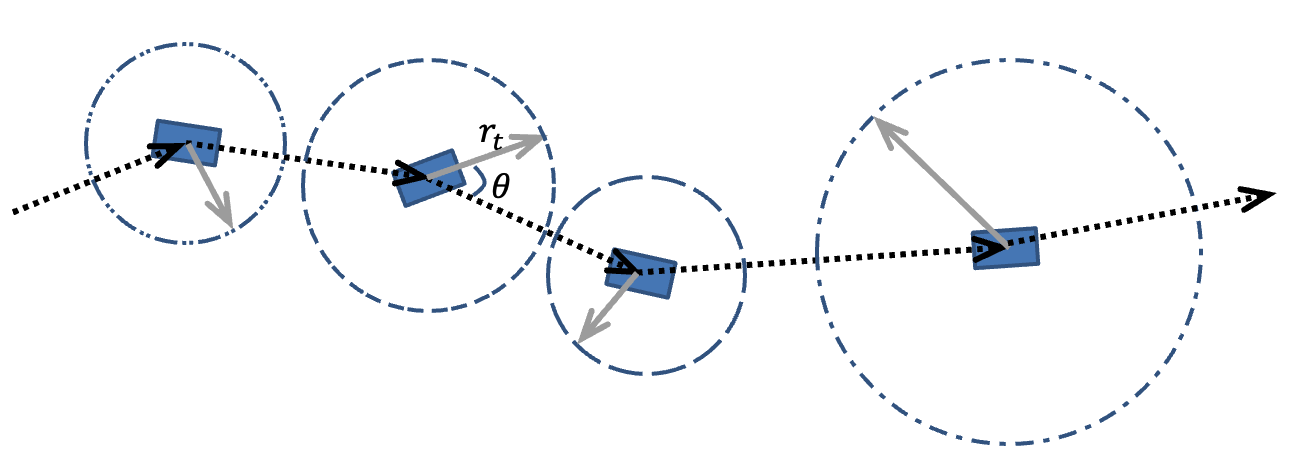}
         \caption{Sampled points for exploration}
         \label{fig:xy_explore1}
     \end{subfigure}

     \begin{subfigure}{1\textwidth}
         \centering
         \includegraphics[width=.9\linewidth]{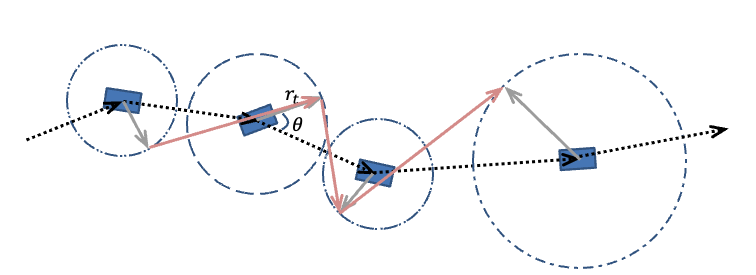}
         \caption{Connected trajectory for exploration}
         \label{fig:xy_explore2}
     \end{subfigure}
     \caption{Exploration scheme in the Cartesian coordinate}
\end{figure}
As is mentioned in Sec. \ref{sec:action}, the action is a planned driving trajectory $(x_1, y_1, \theta_1)$, $(x_2, y_2, \theta_2)$, $(x_3, y_3, \theta_3)$, ..., $(x_T, y_T, \theta_T)$. In the first version of the implementation, the trajectory is represented as the relative displacement from the current point at $t=0$ in the Cartesian coordinate together with the heading information. The RL agent learns through exploration over the optimal policy estimated so far. With this coordinate representation, the exploration is chosen such that each exploration trajectory point is evenly sampled within a circle which centers at the mean of the policy output (Fig. \ref{fig:xy_explore0}) at that timestep with half of the displacement length at that predictive timestamp as the radius (Fig. \ref{fig:xy_explore1}). Then we connect the sampled point at each future timestep as the true trajectory for execution (Fig. \ref{fig:xy_explore2}). \

This method of generating the exploration trajectory is simple to implement. However, it fails to take into account the kinetics of the vehicle and may result in unreasonable trajectories. It can be reflected in high acceleration and jerk required to fulfill the trajectory. Additionally, independent sampling of each point leads to a lack of timing correlation in the trajectories.\

Model Predictive Control (MPC) \cite{GARCIA1989335} was used to re-create dynamics-smooth trajectory from the raw sampled trajectory. However, the correspondence between the anchor points on the two trajectories are hard to be established and the heading information is lost. As a result, the heading weights are never updated, which inpsires us to use the polar coordinate representation.

\subsubsection{Polar Coordinate}
\begin{figure}[h]
     \centering
     \begin{subfigure}{1\textwidth}
         \centering
         \includegraphics[width=.65\linewidth]{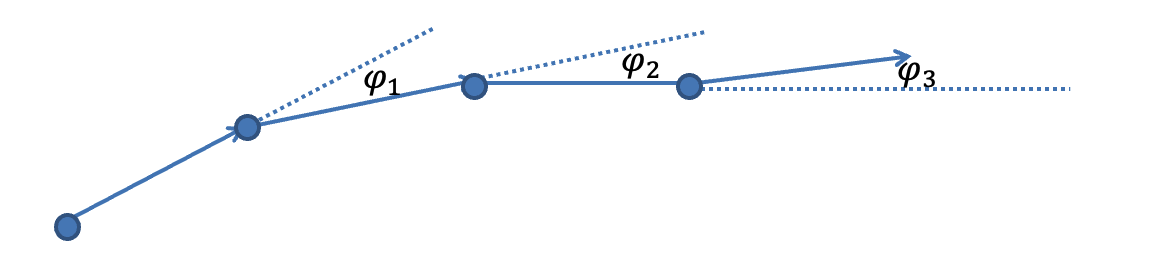}
         \caption{Actions output by the policy network in polar form}
         \label{fig:rphi_explore0}
     \end{subfigure}
     \hfill
     \begin{subfigure}{1\linewidth}
         \centering
         \includegraphics[width=.6\textwidth]{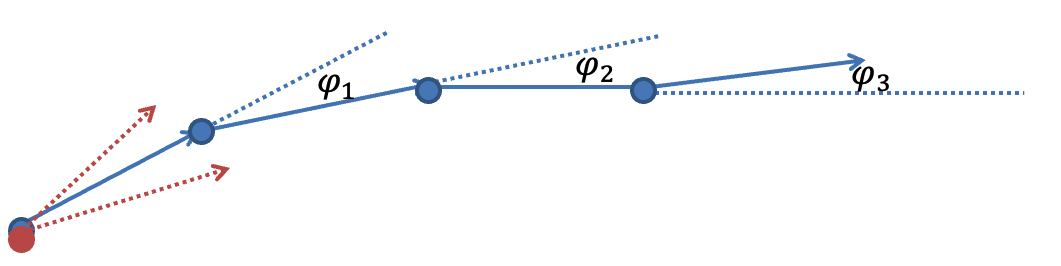}
         \caption{Sampled $\phi_0$ for exploration}
         \label{fig:rphi_explore1}
     \end{subfigure}

     \begin{subfigure}{1\textwidth}
         \centering
         \includegraphics[width=.6\linewidth]{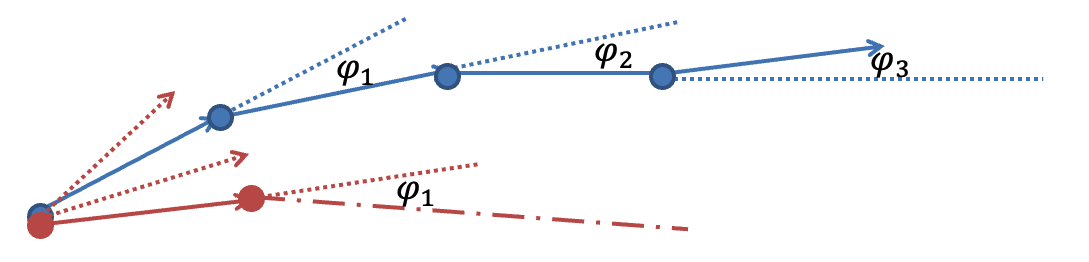}
         \caption{The next sampled $\phi_1$ for exploration}
         \label{fig:rphi_explore2}
     \end{subfigure}
     \begin{subfigure}{1\textwidth}
         \centering
         \includegraphics[width=.6\linewidth]{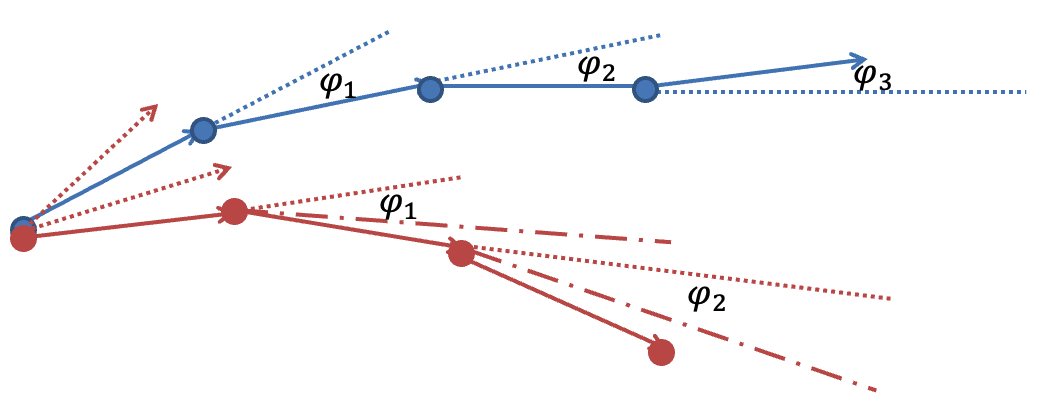}
         \caption{Connected trajectory for exploration}
         \label{fig:rphi_explore3}
     \end{subfigure}
     \caption{Exploration scheme in the Polar coordinate}
\end{figure}
In our final baseline agent implementation, the feature output from the FeatureNet is fed into a recurrent policy network which predicts successive points in the driving trajectory based not only on the input observation but also on the past predictions. To resolve the problem that sampling does not conform to kinetics constraints, we adopt the polar coordinate system $(R, \phi)$, as the trajectory representation. An illustration of the exploration scheme in the polar coordinate system is shown in Fig. \ref{fig:rphi_explore3}.

The vehicle's instantaneous angular speed and acceleration limits can be well represented under the polar coordinate system using $(\Dot{R}, \Dot{\phi})$. The derivative is then approximated by the relative polar coordinate change in a small time interval (sampling interval). The new trajectory $(\delta R_1, \delta \phi_1)$, $(\delta R_2, \delta \phi_2)$, $(\delta R_3, \delta \phi_3)$, ..., $(\delta R_T, \delta \phi_T)$ is then represented by the relative change in polar form from the last point. The exploring point $(\delta R_e, \delta \phi_e)$ is now with its $\delta R_e, \delta \phi_e$ sampled from two Gaussian distributions where the $\delta R, \delta \phi$ are the output from the network as the mean and a controllable variance, respectively. 
\subsubsection{Sampling with Dynamics Constraints}
\begin{figure}[t]
    \centering
    \includegraphics[width=0.6\textwidth]{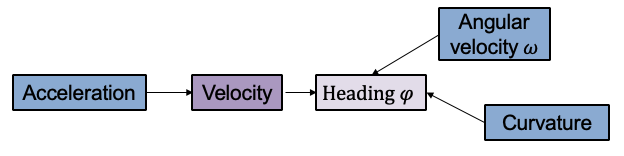}
    \caption{The dynamics constraints among the physical quantities}
    \label{fig:constraints}
\end{figure}

In the exploration above, both $(\delta R_e, \delta \phi_e)$ are sampled according to a Gaussian distribution. However, the feasible execution ranges are limited for both $(\delta R_e, \delta \phi_e)$ in the real situations. Fig. \ref{fig:constraints} demonstrates the causality among the acceleration, velocity, heading, angular velocity, and curvature. The acceleration, the angular velocity, and the curvature are the quantities directly being constrained. They impose the constraints to the velocity and the heading, as the arrows indicate. If we simply truncate the distribution to the desired range, then the gradient of these samples would accumulate to the boundary of the range, resulting in a distorted distribution. 
\begin{figure}[t]
    \centering
    \includegraphics[width=0.6\textwidth]{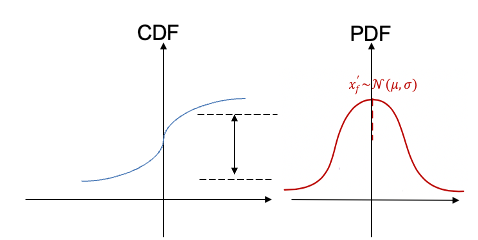}
    \caption{The interval sampling method scheme}
    \label{fig:intervalsp}
\end{figure}

We propose an interval sampling method to tackle the problem. Firstly, we calculated the conditional density function (CDF) of the original distribution, as shown in Fig. \ref{fig:intervalsp}. Then, we limited the sampling range in the CDF domain, as shown with the dashed lines. This range was then mapped to the probability density function (PDF) domain to obtain the real samples. By this means, the original distribution is preserved while the samples are also sampled according to the dynamics constraints.

\subsubsection{Other Exploration Methods}
Other than sampling around the mean value output from the network, we proposed two exploration heuristics, named the exploring tree and the routine fuzzing method. These methods synthesize the data in the form of perturbation to the network output.

The exploring tree method replays the scene and adds exploration noises from multiple timestamps before the failure happens. The routine fuzzing method increases the set diversity by choosing massive random but allowable locations for either the ego vehicle or the surrounding objects, and then synthesizes new scenes. To use these exploration techniques, the vehicle will first be controlled using the action directly output from the policy network for evaluation. Then depending on the evaluation performance, we use the routine fuzzing method on the good ones and apply the exploring tree method to the bad cases around the failure locations.

\subsection{Trajectory Generator}
\begin{figure}[t]
     \centering
     \begin{subfigure}{0.45\textwidth}
         \centering
         \includegraphics[width=.9\linewidth]{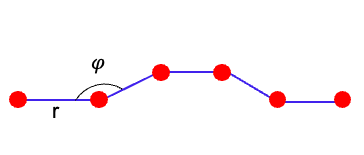}
         \caption{Trajectory with sparse anchor points}
         \label{fig:sparse}
     \end{subfigure}
     \hfill
     \begin{subfigure}{0.45\linewidth}
         \centering
         \includegraphics[width=.9\textwidth]{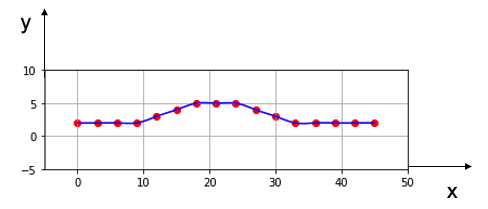}
         \caption{Trajectory with dense anchor points}
         \label{fig:dense}
     \end{subfigure}
     \caption{Input (a) and output (b) of the trajectory generator}
     \label{fig:tra-gen}
\end{figure}

To ensure the trajectory output from the policy network is feasible for the vehicle and a comfortable experience for the passengers, we proposed a trajectory generator that smooths the output trajectory as is shown in Fig. \ref{fig:tra-gen}. The trajectory with sparse anchor points in Fig. \ref{fig:sparse} is transferred to the trajectory with dense anchor points in Fig. \ref{fig:dense}, with the latter satisfying the kinematic feasibility with improved smoothness. The comfort of the passengers is ensured with the optimization objective to minimize both the lateral and the longitudinal accelerations of the vehicle. We formulated the problem as a optimization problem with its objective function and the constraints shown below:
\begin{equation}
\begin{split}
   \min f(P)&=P^\intercal QP~,\quad
    \text{s.t.}\quad A_{eq}P = b_{eq}~\wedge
     A_{ieq}P \leq b_{ieq}~,
\end{split}
\end{equation}
where $P$ is the parameter matrix of the polynomial function $Y(t)$, $Q$ represents the physical quantity matrix. The constraints $A_{eq}P = b_{eq}$ and $A_{ieq}P \leq b_{ieq}$ encoded the requirements of the dynamics.
The relationship between $f(P)$ and $Y(t)$ is revealed below:
\begin{equation}
    \int_0^T(\dddot{Y}(t))^2 dt\\
    =\sum_{i=1}^k\int_{t_{i-1}}^{t_i} ( \dddot{Y}(t) )^2 dt\\
    =\sum_{i=1}^k P^\intercal Q_iP=P^\intercal QP\\
\end{equation}
where $Y(t)$ is a polynomial representation of the whole trajectory and $\dddot{Y}(t)$ is its third derivative.

\subsection{Behavior Cloning}
To ease the RL training complication at the initial stage where exploration is completely random, drawn on from \cite{ChauffeurNet}, supervised learning (or behavior cloning, BC \cite{torabi2018behavioral}) is used to imitate expert drivers. The open-loop data from the real driving situations are auto-labeled and used for imitation learning. The agent uses the same network structure in Behavior Cloning as in the RL training. The FeatureNet is frozen or at least trained with a small learning rate for stability reasons. The Batch Normalization \cite{pmlr-v37-ioffe15} layer of the FeatureNet is also frozen in the PPO training in case of large KL (Kullback–Leibler divergence) values.

\subsection{Results}
\begin{figure}[h]
     \centering
     \begin{subfigure}{0.45\textwidth}
         \centering
         \includegraphics[width=.95\linewidth]{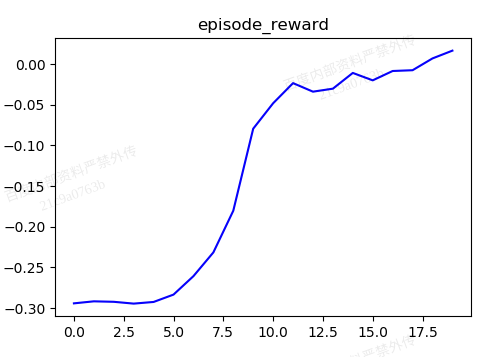}
         \caption{ADS ID: 1011640172567}
         \label{fig:reward1}
     \end{subfigure}
     \hfill
     \begin{subfigure}{0.45\linewidth}
         \centering
         \includegraphics[width=.95\textwidth]{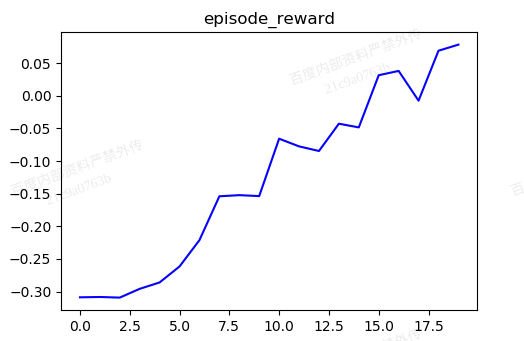}
         \caption{ADS ID: 1021640172567}
         \label{fig:reward2}
     \end{subfigure}

     \begin{subfigure}{0.45\textwidth}
         \centering
         \includegraphics[width=.95\linewidth]{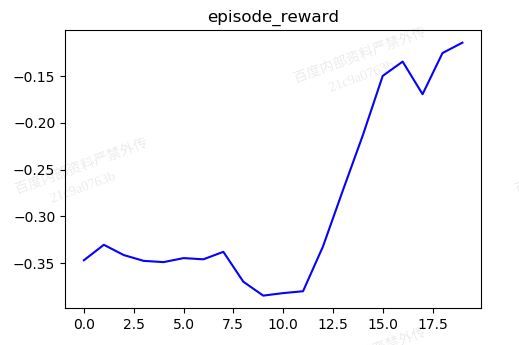}
         \caption{ADS ID: 1031640172567}
         \label{fig:reward3}
     \end{subfigure}
     \hfill
     \begin{subfigure}{0.45\linewidth}
         \centering
         \includegraphics[width=.95\textwidth]{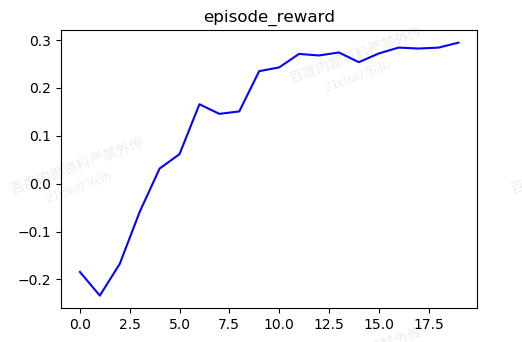}
         \caption{ADS ID: 1041640172567}
         \label{fig:reward4}
     \end{subfigure}
     \caption{Case study curves of accumulated rewards.}\label{fig:reward}
\end{figure}

We pack one or a batch of ADS (Autonomous Driving Scenario) as a whole for training. Usually, each ADS in one training task will be run $N$ times with $N$ ranging from 100-300 and different random exploration strategies in a simulation job. And then batch training will be carried out. Finally, the models are chosen for the next round of iteration. For small-scale scene sets ($<100$ ADS), convergence effects are achieved in around 10-20 rounds of iteration, and more than 90 percent of the scenes reach the highest reward.
While achieving superior performance is not the major purpose of this paper, we leave out the detailed effects, hyperparameters and process analysis. We show the accumulated reward (total return) curves of some cases in Fig. \ref{fig:reward}. The horizontal and the vertical axes represent the training steps and the accumulated rewards, respectively. The plots reveal a steady improvement in the return and are reproducible. The videos of these case studies will be dynamically updated on our official website\footnote{Link: \url{ https://studio.apollo.auto/Apollo-Homepage-Document/Apollo\_Studio}}. We may open a leaderboard and provide evaluation statistical tables as part of our future work.

\section{Conclusion}
This paper introduces ApolloRL as an open platform for deep reinforcement learning research in real world autonomouse driving. The platform contains a reinforcement learning environment and highly integrated distributed simulation, evaluation and training components. We pretrain our baseline agent with supervised learning from expert driver logs and then use reinforcement learning to refine it in the virtual environment. We mainly regard ADS as a way of unit testing. That is to say, the RL agent should be able to complete one or a batch of ADS within a short time (20-30 seconds). By completing adequate complex and long ADSs, an agent may eventually solve a wide range of scenarios or even all of them.

We observe that with our improvement made based on state-of-the-art algorithms, some of the performance of ADS can be improved over the baseline approaches. Some of the others are even better than our on-road autonomous driving system. However, the agent cannot directly replace our system in the actual road test, let alone experienced drivers. In terms of training scales, we have tested dozens of ADS, and the overall training effect has achieved the goal in a large proportion, but there is still a big gap compared to the scale of modern supervised learning. In addition, about algorithm efficiency, we believe there is a large room for improvement, especially for thousands or more ADSs. We hope our work could inspired more exiciting future research in reinforcement learning for autonomous driving.

\section{Future Plan}

We aim to build ApolloRL and an active community around it, \textit{i.e.}, to interact with users and continuously listening to the users’ needs and experiences. We hope to integrate the collective wisdom of reinforcement learning researchers based on this platform and gradually replace the on-road autonomous driving system in more scenarios. The objective of the platform is to develop an agent close to or even beyond the level of human drivers in the real world. 
In the future, we plan to open up more scenarios and more user defined capabilities. In terms of the approaches, more algorithms, models and exploration methods will be released. Another interesting and probably equally important direction is to provide training and evaluation for offline reinforcement learning. In terms of simulation, smart agents and smart cities are considered to better support multi-agent RL research in autonomous driving.

\section*{Acknowledgement}

This system is the result of the work of the Baidu RL team. We list the contributions of team members here: Fei Gao, co-founder of the RL platform; led technical aspects of the RL research in the Planning and Control team (PNC) and set research projects; Peng Geng, member of the PNC team; contributed to the RL algorithms implementation, the trajectory generator, and the automated training pipeline;
Jiaqi Guo, member of the PNC team, contributed to the RL algorithms and model implementation;
Yuan Liu, member of the PNC team, built the interface to the simulation platform and contributed to the RL exploration method development; Dingfeng Guo, member of the PNC team; developed the trajectory generator; Yabo Su, member of the PNC team, contributed to the model development; Jie Zhou, co-founder of the RL platform, technical leader of the RL platform and architect of the simulation platform; Xiao Wei, Jin Li, Xu Liu contributed to the development of the RL platform; the optimization and acceleration of the training process; interface to the simulation and evaluation, training of small batches, etc; Wenzhe Zhang supported the framework for simulation; Jing Wang, technical leader in simulation evaluation and simulation models; Shuai Lu supported the simulation of the agent and the routing fuzzing running method; Dongyi Zhou contributed to the optimization of the simulation metrics; Hang Wu supported the interface to the simulation platform.

\bibliographystyle{unsrt}
\bibliography{ref.bib}

\begin{thebibliography}{10}

\bibitem{av}
Ekim Yurtsever, Jacob Lambert, Alexander Carballo, and Kazuya Takeda.
\newblock A survey of autonomous driving: Common practices and emerging
  technologies.
\newblock {\em IEEE Access}, 8:58443–58469, 2020.

\bibitem{av-safety}
Safety first for autonomous driving, 2019.

\bibitem{Hancock7684}
P.~A. Hancock, Illah Nourbakhsh, and Jack Stewart.
\newblock On the future of transportation in an era of automated and autonomous
  vehicles.
\newblock {\em Proceedings of the National Academy of Sciences},
  116(16):7684--7691, 2019.

\bibitem{Steck2018HowAD}
Felix Steck, Viktoriya Kolarova, Francisco~J. Bahamonde-Birke, Stefan Trommer,
  and Barbara Lenz.
\newblock How autonomous driving may affect the value of travel time savings
  for commuting.
\newblock {\em Transportation Research Record}, 2672:11 -- 20, 2018.

\bibitem{SilverHuangEtAl16nature}
David Silver, Aja Huang, Chris~J. Maddison, Arthur Guez, Laurent Sifre, George
  van~den Driessche, Julian Schrittwieser, Ioannis Antonoglou, Veda
  Panneershelvam, Marc Lanctot, Sander Dieleman, Dominik Grewe, John Nham, Nal
  Kalchbrenner, Ilya Sutskever, Timothy Lillicrap, Madeleine Leach, Koray
  Kavukcuoglu, Thore Graepel, and Demis Hassabis.
\newblock Mastering the game of {Go} with deep neural networks and tree search.
\newblock {\em Nature}, 529(7587):484--489, Jan 2016.

\bibitem{dota2}
Christopher Berner, Greg Brockman, Brooke Chan, Vicki Cheung, Przemyslaw
  Debiak, Christy Dennison, David Farhi, Quirin Fischer, Shariq Hashme,
  Christopher Hesse, Rafal J{\'{o}}zefowicz, Scott Gray, Catherine Olsson,
  Jakub Pachocki, Michael Petrov, Henrique~Pond{\'{e}} de~Oliveira~Pinto,
  Jonathan Raiman, Tim Salimans, Jeremy Schlatter, Jonas Schneider, Szymon
  Sidor, Ilya Sutskever, Jie Tang, Filip Wolski, and Susan Zhang.
\newblock Dota 2 with large scale deep reinforcement learning.
\newblock {\em CoRR}, abs/1912.06680, 2019.

\bibitem{muzero}
Julian Schrittwieser, Ioannis Antonoglou, Thomas Hubert, Karen Simonyan,
  Laurent Sifre, Simon Schmitt, Arthur Guez, Edward Lockhart, Demis Hassabis,
  Thore Graepel, and et~al.
\newblock Mastering atari, go, chess and shogi by planning with a learned
  model.
\newblock {\em Nature}, 588(7839):604–609, Dec 2020.

\bibitem{openai2019learning}
OpenAI, Marcin Andrychowicz, Bowen Baker, Maciek Chociej, Rafal Jozefowicz, Bob
  McGrew, Jakub Pachocki, Arthur Petron, Matthias Plappert, Glenn Powell, Alex
  Ray, Jonas Schneider, Szymon Sidor, Josh Tobin, Peter Welinder, Lilian Weng,
  and Wojciech Zaremba.
\newblock Learning dexterous in-hand manipulation, 2019.

\bibitem{mirowski2018learning}
Piotr Mirowski, Matthew~Koichi Grimes, Mateusz Malinowski, Karl~Moritz Hermann,
  Keith Anderson, Denis Teplyashin, Karen Simonyan, Koray Kavukcuoglu, Andrew
  Zisserman, and Raia Hadsell.
\newblock Learning to navigate in cities without a map.
\newblock In {\em Neural Information Processing Systems (NeurIPS)}, 2018.

\bibitem{streetair_iccv2019}
Ang Li, Huiyi Hu, Piotr Mirowski, and Mehrdad Farajtabar.
\newblock Cross-view policy learning for street navigation.
\newblock In {\em International Conference on Computer Vision (ICCV)}, 2019.

\bibitem{ale}
Marc~G. Bellemare, Yavar Naddaf, Joel Veness, and Michael Bowling.
\newblock The arcade learning environment: An evaluation platform for general
  agents.
\newblock {\em J. Artif. Int. Res.}, 47(1):253–279, May 2013.

\bibitem{starcraft}
Oriol Vinyals, Igor Babuschkin, Wojciech~M. Czarnecki, Micha{\"e}l Mathieu,
  Andrew Dudzik, Junyoung Chung, David~H. Choi, Richard Powell, Timo Ewalds,
  Petko Georgiev, Junhyuk Oh, Dan Horgan, Manuel Kroiss, Ivo Danihelka, Aja
  Huang, Laurent Sifre, Trevor Cai, John~P. Agapiou, Max Jaderberg,
  Alexander~S. Vezhnevets, R{\'e}mi Leblond, Tobias Pohlen, Valentin Dalibard,
  David Budden, Yury Sulsky, James Molloy, Tom~L. Paine, Caglar Gulcehre, Ziyu
  Wang, Tobias Pfaff, Yuhuai Wu, Roman Ring, Dani Yogatama, Dario W{\"u}nsch,
  Katrina McKinney, Oliver Smith, Tom Schaul, Timothy Lillicrap, Koray
  Kavukcuoglu, Demis Hassabis, Chris Apps, and David Silver.
\newblock Grandmaster level in starcraft ii using multi-agent reinforcement
  learning.
\newblock {\em Nature}, 575(7782):350--354, 2019.

\bibitem{deepmindlab}
Charles Beattie, Joel~Z. Leibo, Denis Teplyashin, Tom Ward, Marcus Wainwright,
  Heinrich Küttler, Andrew Lefrancq, Simon Green, Víctor Valdés, Amir Sadik,
  Julian Schrittwieser, Keith Anderson, Sarah York, Max Cant, Adam Cain, Adrian
  Bolton, Stephen Gaffney, Helen King, Demis Hassabis, Shane Legg, and Stig
  Petersen.
\newblock Deepmind lab, 2016.

\bibitem{openaigym}
Greg Brockman, Vicki Cheung, Ludwig Pettersson, Jonas Schneider, John Schulman,
  Jie Tang, and Wojciech Zaremba.
\newblock Openai gym, 2016.

\bibitem{torcs}
Bernhard Wymann, Christos Dimitrakakisy, Andrew Sumnery, and Christophe
  Guionneauz.
\newblock Torcs: The open racing car simulator, 2015.

\bibitem{carla}
Alexey Dosovitskiy, German Ros, Felipe Codevilla, Antonio Lopez, and Vladlen
  Koltun.
\newblock Carla: An open urban driving simulator, 2017.

\bibitem{sumo}
Pablo~Alvarez Lopez, Michael Behrisch, Laura Bieker-Walz, Jakob Erdmann,
  Yun-Pang Flötteröd, Robert Hilbrich, Leonhard Lücken, Johannes Rummel,
  Peter Wagner, and Evamarie Wiessner.
\newblock Microscopic traffic simulation using sumo.
\newblock In {\em 2018 21st International Conference on Intelligent
  Transportation Systems (ITSC)}, pages 2575--2582, 2018.

\bibitem{smarts}
Ming Zhou, Jun Luo, Julian Villella, Yaodong Yang, David Rusu, Jiayu Miao,
  Weinan Zhang, Montgomery Alban, Iman Fadakar, Zheng Chen, Aurora~Chongxi
  Huang, Ying Wen, Kimia Hassanzadeh, Daniel Graves, Dong Chen, Zhengbang Zhu,
  Nhat Nguyen, Mohamed Elsayed, Kun Shao, Sanjeevan Ahilan, Baokuan Zhang,
  Jiannan Wu, Zhengang Fu, Kasra Rezaee, Peyman Yadmellat, Mohsen Rohani,
  Nicolas~Perez Nieves, Yihan Ni, Seyedershad Banijamali, Alexander~Cowen
  Rivers, Zheng Tian, Daniel Palenicek, Haitham bou Ammar, Hongbo Zhang, Wulong
  Liu, Jianye Hao, and Jun Wang.
\newblock Smarts: Scalable multi-agent reinforcement learning training school
  for autonomous driving, 2020.

\bibitem{driving}
Parth Kothari, Christian Perone, Luca Bergamini, Alexandre Alahi, and Peter
  Ondruska.
\newblock Drivergym: Democratising reinforcement learning for autonomous
  driving, 2021.

\bibitem{ChauffeurNet}
Mayank Bansal, Alex Krizhevsky, and Abhijit Ogale.
\newblock Chauffeurnet: Learning to drive by imitating the best and
  synthesizing the worst, 2018.

\bibitem{pixel}
Yuchi Tian, Kexin Pei, Suman Jana, and Baishakhi Ray.
\newblock Deeptest: Automated testing of deep-neural-network-driven autonomous
  cars, 2018.

\bibitem{ppo}
John Schulman, Filip Wolski, Prafulla Dhariwal, Alec Radford, and Oleg Klimov.
\newblock Proximal policy optimization algorithms, 2017.

\bibitem{sac}
Tuomas Haarnoja, Aurick Zhou, Pieter Abbeel, and Sergey Levine.
\newblock Soft actor-critic: Off-policy maximum entropy deep reinforcement
  learning with a stochastic actor, 2018.

\bibitem{GARCIA1989335}
Carlos~E. Garc{\'\i}a, David~M. Prett, and Manfred Morari.
\newblock Model predictive control: Theory and practice---a survey.
\newblock {\em Automatica}, 25(3):335--348, 1989.

\bibitem{torabi2018behavioral}
Faraz Torabi, Garrett Warnell, and Peter Stone.
\newblock Behavioral cloning from observation, 2018.

\bibitem{pmlr-v37-ioffe15}
Sergey Ioffe and Christian Szegedy.
\newblock Batch normalization: Accelerating deep network training by reducing
  internal covariate shift.
\newblock In Francis Bach and David Blei, editors, {\em Proceedings of the 32nd
  International Conference on Machine Learning}, volume~37 of {\em Proceedings
  of Machine Learning Research}, pages 448--456, Lille, France, 07--09 Jul
  2015. PMLR.

\end{thebibliography}

\end{document}